\documentclass[conference]{IEEEtran}
\IEEEoverridecommandlockouts
\usepackage{cite,hyperref,bm,amsmath,amssymb,amsfonts,algorithmic,algorithm,graphicx,textcomp,xcolor,comment,subfiles,caption,subcaption,gensymb,array, authblk}
\usepackage{tikz,pgfplots,pgfplotstable}

\graphicspath{{images/}}
\begin{document}
\title{Hierarchical Multi-Agent Reinforcement Learning for Air Combat Maneuvering}

\author[1]{Ardian Selmonaj}
\author[1]{Oleg Szehr}
\author[1]{Giacomo Del Rio}
\author[1]{Alessandro Antonucci}
\author[2]{\authorcr Adrian Schneider}
\author[2]{Michael Rüegsegger}

\affil[1]{Istituto Dalle Molle di Studi sull'Intelligenza Artificiale (IDSIA), USI-SUPSI, Lugano (Switzerland)}
\affil[2]{Armasuisse Science + Technology, Thun (Switzerland)}

\maketitle
\begin{abstract}
The application of artificial intelligence to simulate air-to-air combat scenarios is attracting increasing attention. To date the high-dimensional state and action spaces, the high complexity of situation information (such as imperfect and filtered information, stochasticity, incomplete knowledge about mission targets) and the nonlinear flight dynamics pose significant challenges for accurate air combat decision-making. These challenges are exacerbated when multiple heterogeneous agents are involved. We propose a hierarchical multi-agent reinforcement learning framework for air-to-air combat with multiple heterogeneous agents. In our framework, the decision-making process is divided into two stages of abstraction, where heterogeneous low-level policies control the action of individual units, and a high-level commander policy issues macro commands given the overall mission targets. Low-level policies are trained for accurate unit combat control. Their training is organized in a learning curriculum with increasingly complex training scenarios and league-based self-play. The commander policy is trained on mission targets given pre-trained low-level policies. The empirical validation advocates the advantages of our design choices.
\end{abstract}
\begin{IEEEkeywords}
Hierarchical Multi-Agent Reinforcement Learning, Heterogeneous Agents, Curriculum Learning, Air Combat.
\end{IEEEkeywords}
\section{Introduction}
\label{intro}
In defense area, complex air-to-air combat scenarios simulation requires simultaneous real-time control of individual units (troop level) and global mission planning (commander level). \emph{Deep Reinforcement Learning} (DRL) has achieved super-human level in various environments, ranging from discrete perfect information scenarios (such as games like Chess and Go) to real-time continuous control and strategic decision-making scenarios with imperfect information (typical of modern war games). However, conventional DRL ignores the structural requirements typical of real-world combat scenarios, where decision-making authority is organized hierarchically. It is crucial for real-world operations that low-level combat decisions (such as fire/duck) are made by individual units and executed at low latency, while abstract mission planning decisions (such as conquer and hold coordinates) at higher hierarchy levels take account of information from all available units. An example is the guidance of drones in modern warfare, where individual units act autonomously even without connection to a centralized intelligent instance. This information abstraction motivates the investigation of \emph{Multi-Agent Deep Reinforcement Learning} (MARL) techniques for creating artificial agents in a realistic simulation environment. In MARL systems, the hierarchical splitting of mixed planning and control tasks can be achieved by incorporating dedicated algorithms at varying levels of abstraction. This allows each agent to control itself in a decentralized manner while providing sufficient flexibility for the emergence of targeted group behavior. 

\subsection{Contributions}
\begin{enumerate}
\item Considering low latency as crucial for DRL problems, we develop a lightweight simulation platform suitable for fast simulation of agent dynamics and interactions.
\item We employ a hierarchical framework for simultaneous planning and control to solve the overall decision-making problem for air-to-air combat scenarios. 
\item We realize a fictitious self-play mechanism through curriculum learning with increasing levels of complexity to improve combat performance as learning proceeds.
\item We develop a sophisticated neural network architecture composed of recurrent and attention units. Coordination is achieved without an explicit communication channel. 
\end{enumerate}

\subsection{Outline}
Sect.~\ref{sec:related} summarizes previous contributions and describes how our work extends the existing literature. Sect.~\ref{sec:method} details air-to-air engagement scenarios and describes our framework. Our experimental findings are presented in Sect.~\ref{sec:experiments}, while our conclusions and possible future works are discussed in  Sect.~\ref{sec:conclusions}.

\section{Related Work}\label{sec:related}
Aerial combat tactics have been discussed extensively in the literature, with a significant portion of research dedicated to the study of engagements with small numbers of units (one to two). Research on small engagements typically focuses on \emph{control}, i.e., it examines how the maneuvering of individual units impacts the overall engagement outcome. A frequent focus lies on achieving an advantage against the opponent: in this position, it is possible to fire at the opponent with little risk of return fire~\cite{shawFight}. Popular methods include expert systems \cite{burgin_1,burgin_2,jones_1,jones_2}, control laws for pursuit and/or evasion \cite{Eklund,Virtanen2004,Virtanen2006,You2014}, game-theory \cite{Jarmark1984,Merz1985,Greenwood1992}, but also machine learning \cite{McGrew2010,Ma2018,Day2018,Vlahov2018} and hybrid approaches \cite{Smith2000,Smith2000_,Smith2001,Smith2002,Smith2004,Toubman2015,Toubman2016}. Classical research about larger-scale engagements focuses on weapon-target assignment \cite{optimumTargetAssignment} and \cite{shawFight},  human-pilot-like decision-making \cite{Tidhar1998}, and high-level engagement tactical decisions \cite{dayThesis}, i.e., on \emph{planning}. \emph{Reinforcement learning} (RL) techniques gained increasing interest in this context. \cite{rl_combat6} train a Recurrent \emph{Deep $Q$-Network} (DQN) algorithm \cite{mnih2013playing} and employ a situation evaluation function to revise the decision-making system. Other approaches use \emph{deep deterministic policy gradient} (DDPG) \cite{rl_combat0, rl_combat1} or A3C \cite{rl_combat7}. \emph{Cascade learning} approaches that gradually increase combat complexity are discussed in \cite{rl_combat3} and \cite{rl_combat4}. In \cite{rl_combat5}, the combat strategy is learned through a league system to prevent the policy from circling around poor local optima.

MARL is currently thriving \cite{Gronauer2021}. In the study of emergent behavior and complexity from coordinated agents, the introduction of centralized training decentralized execution actor-critic methods, such as the \emph{Multi-Agent Deep Deterministic Policy Gradient} algorithm (MADDPG) \cite{CTDE}, has been a milestone \cite{CTDE,myAss,myAss_2}. Such methods train actor policies with critics having access to information of all other agents. However, they are not structured to account for the hierarchies present in real-world operations and emerging phenomena such as agent attrition (exit learning process). \cite{marl1} uses MADDPG combined with potential-based reward shaping \cite{potential_reward}. A maneuver strategy for \emph{Unmanned Aerial Vehicles} (UAV) swarms is developed in~\cite{marl2} using MADDPG, but the discussion is limited to one-to-one or multi-to-one combat. \cite{marl4} and \cite{marl5} use attention based neural networks. The former uses a two-stage attention mechanism for coordination. In the latter, the attention layers calculate the relative importance of surrounding aircraft, where opponents are purely controlled by scripts. 

On the other side stands the concept of \emph{Hierarchical Reinforcement Learning} (HRL), which divides the overall task into sub-tasks~\cite{hier_def_barto}. In HRL, training is organized in nested loops. Inner loop training controls the aircraft, while the outer trains a super-policy for guidance and coordination of individual agents. HRL has been applied in the context of air-to-air combat in \cite{hier1} and \cite{hier2}.

There appears to be little research in air-to-air combat by combining HRL and MARL. A Hierarchical MARL approach to handle variable-size formations is proposed in \cite{marl3}. The authors employ an attention mechanism and self-play with a DQN high-level policy trained with QMIX. An approach similar to the one presented in this paper, focusing on heterogeneous agents of two types, was explored in \cite{marl_hier_hetero}. The high-level target allocation agents are trained using DQN, and the low-level cooperative attacking agents are based on \emph{independent asynchronous proximal policy optimization} (IAPPO). However, they follow the goal of \emph{suppression of enemy air defense} (SEAD). SEAD aims to gain air superiority by targeting and disrupting the enemy's ability to detect and engage friendly aircraft. Unlike the concept of SEAD, which focuses on neutralizing enemy air defense systems, dogfighting is centered around engaging and defeating enemy aircraft in direct air-to-air combat. This article investigates air-to-air combat scenarios for coordinated dogfighting with heterogeneous agents and hierarchical MARL in a cascaded league-play training scheme. To our knowledge, this setup has yet to occur in publications for this kind of application.

\section{Method}\label{sec:method}
\subsection{Aircraft Dynamics}\label{aircraft_dynamics}
We base our modeling on the dynamics of the \emph{Dassault Rafale} fighter aircraft.\footnote{ \href{https://www.dassault-aviation.com/en/defense/rafale}{dassault-aviation.com/en/defense/rafale}.} We focus on hierarchical coordination of multiple heterogeneous agents in 2D (assuming a constant altitude of our aircraft). There are \emph{beyond} and \emph{within visual range} air combat scenarios~\cite{wvr}, where we focus on the latter in this article. Since real-world combat scenarios frequently involve different types of aircraft, we add a modified version of the Rafale aircraft with different dynamics: The original aircraft (AC1) is more agile and equipped with rockets, while the modified type (AC2) has no rockets but longer cannon range. The dynamics of AC1 and AC2 are characterized as:
\begin{itemize}
\item angular velocity [$deg/s$]: $\omega_{AC1}\in[0,5]$, $\omega_{AC2}\in[0,3.5]$;
\item speed [knots]: $v_{AC1}\in [100,900]$, $v_{AC2}\in [100,600]$;
\item conical \emph{weapon engagement zone} (WEZ): \newline
angle [$deg$]: $\omega_{WEZ,AC1}\in[0,10]$, $\omega_{WEZ,AC2}\in[0,7]$, range [$km$]: $d_{a,AC1}\in[0,2]$, $d_{a,AC2}\in[0,4.5]$;
\item hit probability $p_{hit,AC1}=0.70$ and $p_{hit,AC2}=0.85$.
\end{itemize}

\begin{figure}[htbp!]
\centerline{\includegraphics[scale=0.13]{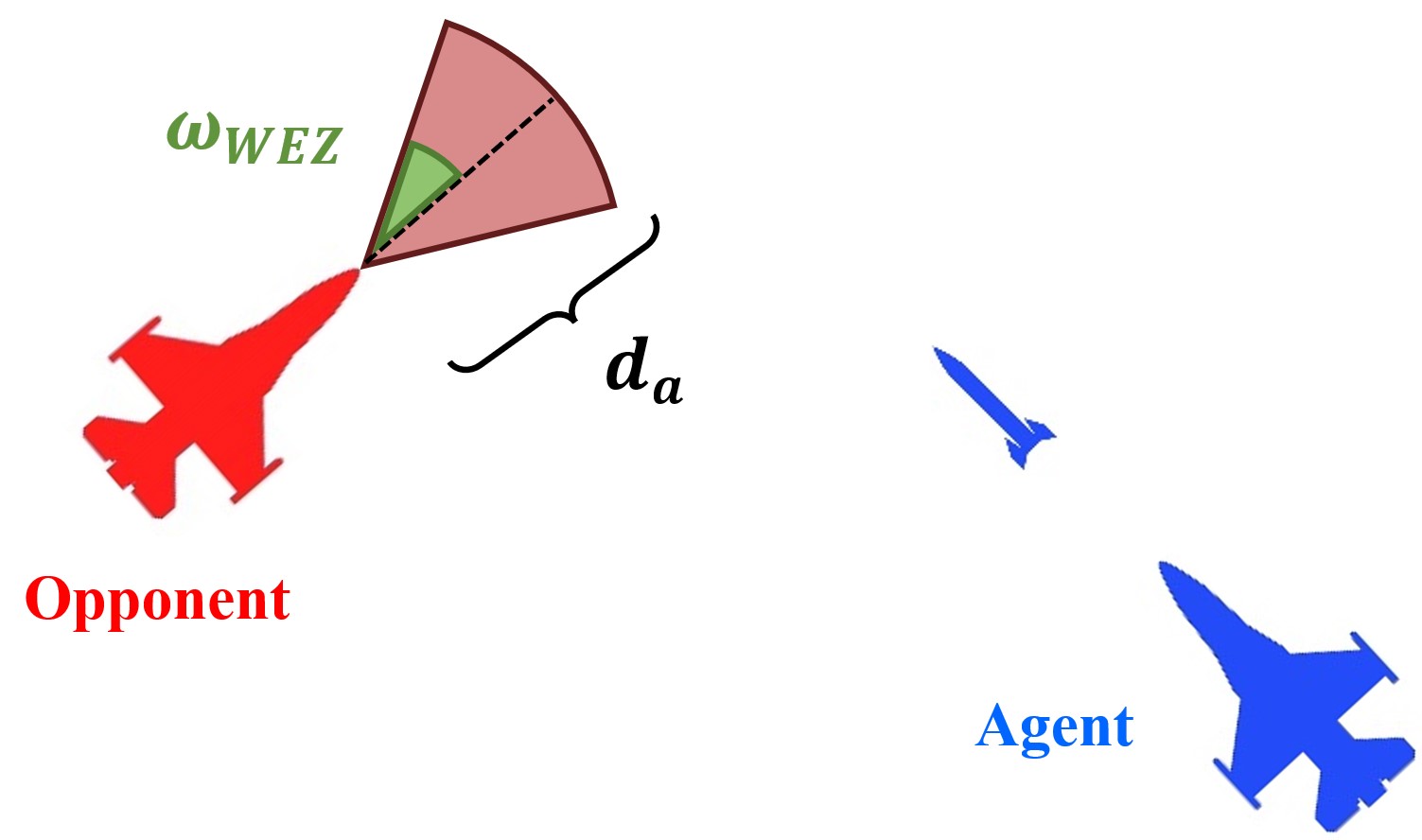}}
\caption{Aircraft attacking mechanisms.}
\label{fig:aircrafts_attack}
\end{figure}

\subsection{Multi-Agent Reinforcement Learning}\label{MARL_def}
RL is used to solve sequential decision-making problems. Agents interact with an environment to learn a behavior that is evaluated based on a reward function $r_t(s_t,a_t, s_{t+1})$. The goal of the agent is to maximize the cumulative reward $\sum_t r_t(s_t, a_t, s_{t+1})$. In MARL, there are multiple agents interacting in a cooperative or competing fashion, or both. The decision function, called policy $\pi(a_t|s_t)$, maps states to a distribution over actions. We model the interactions of individual agents as a \emph{partially-observable Markov game} (POMG) defined by a tuple $(\mathcal{S}, \mathcal{O}, \mathcal{A}_1, \ldots, \mathcal{A}_N, P, R_1, \ldots, R_N, \gamma)$, where: $\mathcal{S}$ is the state-space representing possible configurations of the environment, $\mathcal{O} \subset \mathcal{S}$ is the set of observations, $\mathcal{A}_i$ is the set of actions for player $i$, $P(s'|s, a_1, \ldots, a_N)$ represents the dynamics of the environment and specifies the probability of transitioning to state $s'$ when players take actions $a_1, \ldots, a_N$ in state $s$, $R_i(s, a_1, \ldots, a_N,s')$ defines the immediate reward for player $i$ when the system transitions from state $s$ to state $s'$ with players taking actions $a_1,\ldots,a_N$. 

We adopt a \emph{Centralized Training and Decentralized Execution} (CTDE) \cite{CTDE} scheme for training agents. Our modeled POMG with CTDE scheme is used to train low-level control policies, from which we define to have two: a fight policy $\pi_f$ and an escape policy $\pi_e$. Further on, there is a distinct policy for each aircraft type. Overall we have four low-level policies: $[\pi_{f,AC1}, \pi_{f,AC2}, \pi_{e,AC1}, \pi_{e,AC2}]$. Agents of the same type use the same shared policies. Thus all AC1 use $\pi_{f,AC1}$ and $\pi_{e,AC1}$, irrespective on the number of agents, and similarly for AC2. In this way, policies are trained with experiences of all agents of the same type and ensures a coherent behavior.


\subsection{Hierarchical Reinforcement Learning}
\begin{figure}[htbp]
\centerline{\includegraphics[scale=0.09]{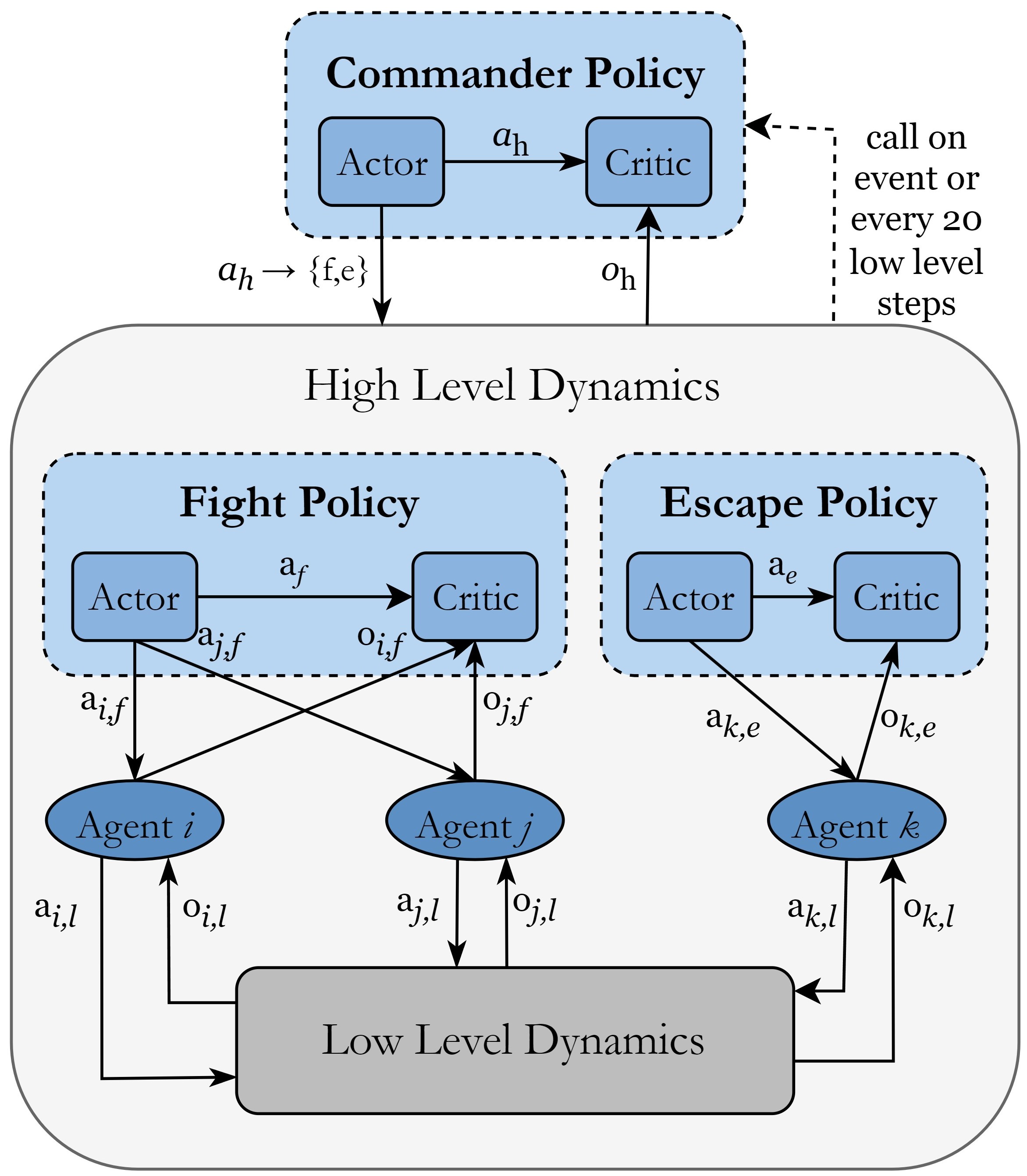}}
\caption{Hierarchy of policies.}
\label{hierarchy_policy}
\end{figure}

HRL employs temporal abstraction by decomposing the overall task into a nested hierarchy of sub-tasks, enhancing efficiency in learning and decision-making~\cite{hier_def_barto}. Abstract commands are issued from higher hierarchy levels to apply a control policy (so-called \emph{option}) for a limited amount of time. Symmetries within a particular (lower) hierarchy level can be exploited by using the same option for different sub-tasks, e.g. controlling similar airplanes. This results in better scalability (reducing the effective dimensions of state and action spaces) and enhances generalization (generating new skills by combining sub-tasks)~\cite{hier_adv}. HRL also fits naturally with the hierarchical structure of defense organizations. Formally, our hierarchical system corresponds to a \emph{partially observable semi-Markov Decision Process} (POSMDP) with options as a tuple $(\mathcal{S}, \mathcal{O}_{s}, \mathcal{A}, R, P, \gamma)$. Similar to the notions of POMG, $\mathcal{S}$ is the state space, $\mathcal{O}_s$ is the set of sub-strategies (options), $\mathcal{A}$ is the action space, $R$ is the reward function and the transition function $P(s', \tau|s,o)$ defines the probability of landing in state $s'$ from state $s$ after $\tau$ time steps when executing $o$. We again use CTDE to train a single high-level commander policy $\pi_h$ to be used for all agents and aircraft types. Fig.~\ref{hierarchy_policy} illustrates the relations between high and low-level policies.

\subsection{Metrics for Air-to-Air Combat}
\begin{figure}[htbp]
\centering
\begin{subfigure}[h]{0.08\textwidth}
\centering
\includegraphics[scale=0.07]{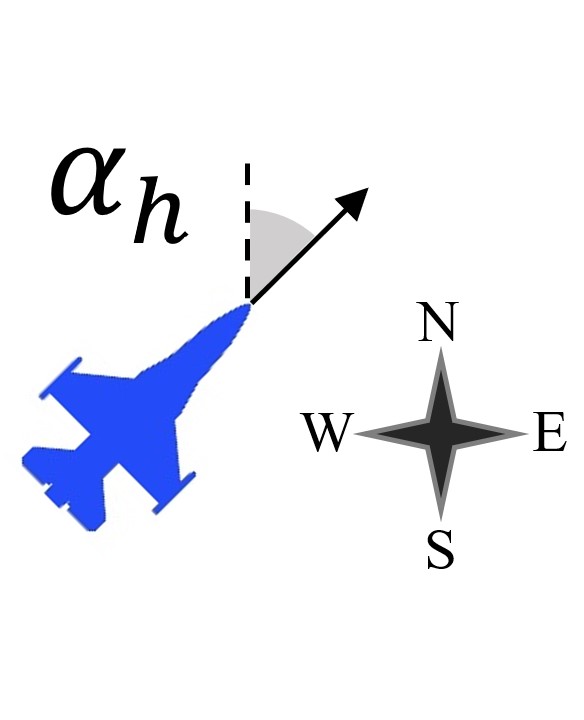}
\caption{}
\end{subfigure}
\begin{subfigure}[h]{0.08\textwidth}
\centering
\includegraphics[scale=0.07]{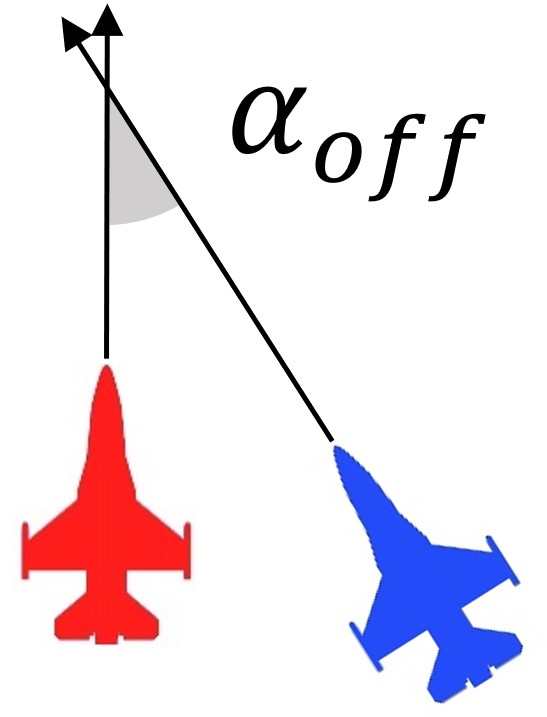} \caption{}
\end{subfigure}
\begin{subfigure}[h]{0.08\textwidth}
\centering
\includegraphics[scale=0.07]{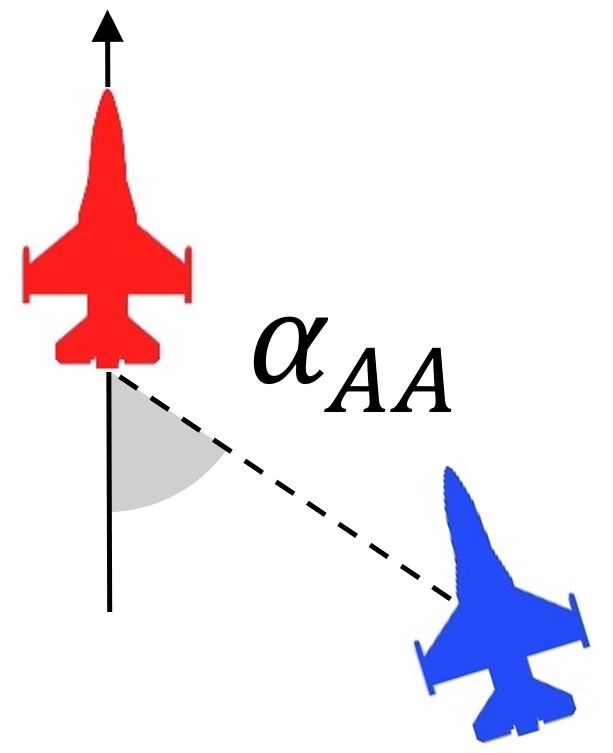} \caption{}
\end{subfigure}
\begin{subfigure}[h]{0.08\textwidth}
\centering
\includegraphics[scale=0.07]{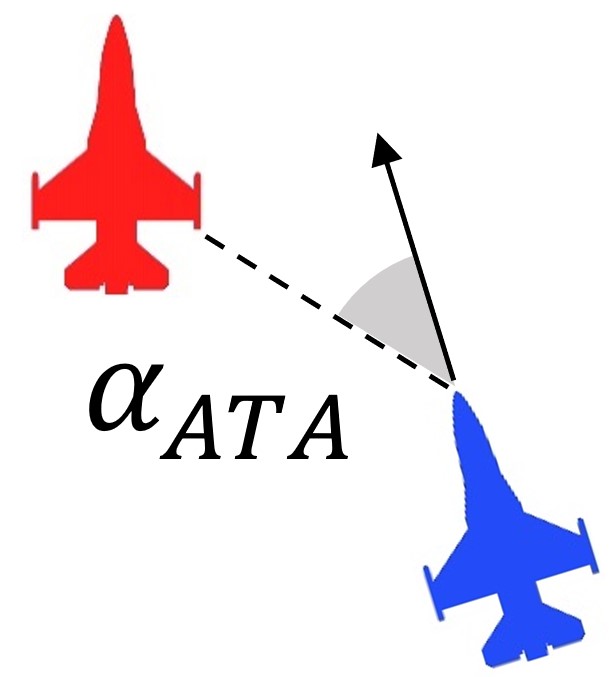}  \caption{}
\end{subfigure}
\begin{subfigure}[h]{0.08\textwidth}
\centering
\includegraphics[scale=0.07]{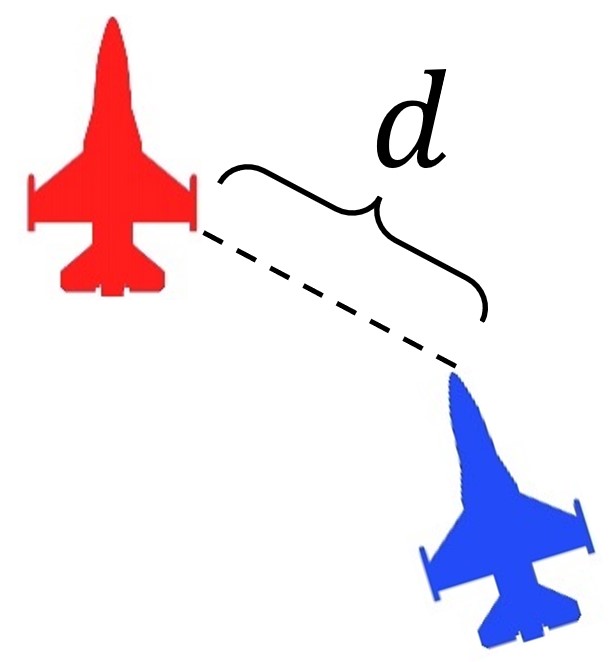}
\caption{}
\end{subfigure}
\caption{Aircraft metrics: heading (a), heading off (b), aspect angle (c), antenna train angle (d), distance (e).}
\label{ac_metrics}
\end{figure}

We now describe observations, actions and rewards in our hierarchical MARL approach. All observation values are normalized to the range $[0,1]$ and are based on the metrics shown in Fig.~\ref{ac_metrics} Further observations include map position ($x,y$), current speed ($s$), remaining cannon ammunition ($c_1$) and remaining rockets ($c_2$). Indicator ($w$) defines if the next rocket is ready to be fired and ($s_r$) indicates if the aircraft is currently shooting. Subscript $a$ indicates agent, $o$ opponent and $fr$ friendly aircraft (i.e. from the same team). A subscript in a value, e.g., $\alpha_{off,o}$, defines the angle-off w.r.t. to the opponent. Actions of all policies are discrete.

\subsubsection{Fight Policy}
$\pi_f$ can observe its closest opponent and closest friendly aircraft.
\begin{eqnarray*}
o_{t,a} &:=& [x, y, s, \alpha_h, \alpha_{off, o}, \alpha_{AA, o}, \alpha_{ATA, o}, d_o, c_1, \overbrace{c_2, w}^\text{AC1}, s_r]\\
o_{t,o} &:=& [x, y, s, \alpha_h, \alpha_{off, a}, \alpha_{AA, a}, \alpha_{ATA, a}, d_a, s_r]\\
o_{t, fr} &:=& [x, y, s, \alpha_{off, a}, \alpha_{ATA,a}, \alpha_{ATA,fr}, d_a, s_r]\\
o_{t, full} &:=& o_{t, a} || o_{t, o} || o_{t, fr} 
\end{eqnarray*}
The control maneuvers (actions) are:
\begin{itemize}
\item relative heading maneuvers: turn in range [-90\textdegree, 90\textdegree] \newline ($h \in \{-6,\ldots,6\} \xrightarrow{} \alpha_h = -15 \cdot h + \alpha_h$);
\item velocity: mapping of $v$ to velocity ranges of AC1 or AC2 ($v \in \{0,\ldots,8\}$);
\item shooting with cannon: ($c \in \{0,1\}$);
\item shooting with rocket (AC1): ($r \in \{0,1\}$).
\end{itemize}

In air-to-air combat, facing the opponent's tail is a favorable situation for shooting. We therefore define the reward function based on $\alpha_{ATA,a}$ of the opponent to the agent. We further encourage the combat efficiency by incorporating the remaining ammunition ($c_{rem} = c_1 + c_2$):
\begin{equation}
r_{t,k} = \alpha_{ATA,a} + \frac{c_{max}-c_{rem}}{c_{max}} \in [1,2]\,.
\end{equation}
Punishing rewards are given when flying out of environment boundaries $r_{t,b}=-5$ and when destroying a friendly aircraft $r_{t,f} = -2$. There is no per-time-step reward given. The total reward is then: $r_t = r_{t,k} + r_{t,b} + r_{t,f}$.

\subsubsection{Escape Policy}
$\pi_e$ senses two closest opponents and its closest friendly aircraft. The actions remain same as for $\pi_f$.
\begin{eqnarray*}
o_{t,a} &:=& [x, y, s, \alpha_h, c_1, \overbrace{c_2}^\text{AC1}] \\
o_{t,o} &:=& [x, y, s, \alpha_h, \alpha_{off,a}, \alpha_{ATA,a}, \alpha_{ATA,o}, d_a] \\
o_{t, fr} &:=& [x, y, s, \alpha_{h}, \alpha_{ATA,a}, \alpha_{ATA,fr}, d_a] \\
o_{t, full} &:=& o_{t, a} || o_{t, o_1} || o_{t, o_2} || o_{t, fr}
\end{eqnarray*}
The per-time-step reward depends on distances to opponents:
\begin{equation}
r_{t,e}=
\begin{cases}
-0.01 & d<6km\\
+0.01 & d>13km \\
0 & \mathrm{otherwise}
\end{cases}\,.
\end{equation}
The total reward is finally $r_t := r_{t,e} + r_{t,b} + r_{t,f}$.

\subsubsection{Commander Policy}\label{method:commander}
$\pi_h$ is called for every agent separately. The observations are based on three closest opponents and two closest friendly aircraft.
\begin{eqnarray*}
o_{t,a} &:=& [x, y, s, \alpha_h] \\
o_{t,o} &:=& [x, y, s, \alpha_h, \alpha_{AA,a}, \alpha_{AA,o}, \alpha_{ATA,a}, \alpha_{ATA,o}, d_a] \\
o_{t, fr} &:=& [x, y, s, d_a] \\
o_{t, full} &:=& o_{t, a} || o_{t, o_1} || o_{t, o_2} || o_{t, o_3} || o_{t, fr_1} || o_{t, fr_2} 
\end{eqnarray*}
The commander decides the low-level policy to use for each agent. The action set is $a_c \in \{0, 1, 2, 3\}$, where $0$ activates $\pi_e$ and $\pi_f$ otherwise. If $\pi_f$ is activated, the commander action ($1$, $2$ or $3$) determines which of the three observable opponents the agent should attack. The agent then gets the corresponding observation for its low-level policy. In our setup, the commander adapts to only these pre-trained low-level policies. The reward is composed of two parts. First, a killing reward $r_{t,f}=1$ if an agent with its low-level policy killed an opponent and $r_{t,f}=-1$ if the agent got killed. The second part, $r_{t,c}$ should encourage the commander to exploit favorable situations and is defined as follows:
\begin{equation}
r_{t,c}=
\begin{cases}
+0.1 & d_o\!\!<\!\!5km\wedge\alpha_{ATA,o}\!\!<\!\!30^\circ\wedge\alpha_{AA,o}\!\!<\!\!50^\circ\wedge a_c\!\!>\!\!0\\
0 & \mathrm{otherwise}
\end{cases}
\label{fav_sit_reward}
\end{equation}
We also include the out-of-boundary reward $r_{t,b}=-5$. Total reward given to $\pi_h$ is thus $r_t := r_{t,f}+r_{t,c}+r_{t,b}$.

\subsection{Training Structure}\label{method:training_structure}
\begin{figure}[htbp]
\centerline{\includegraphics[scale=0.12]{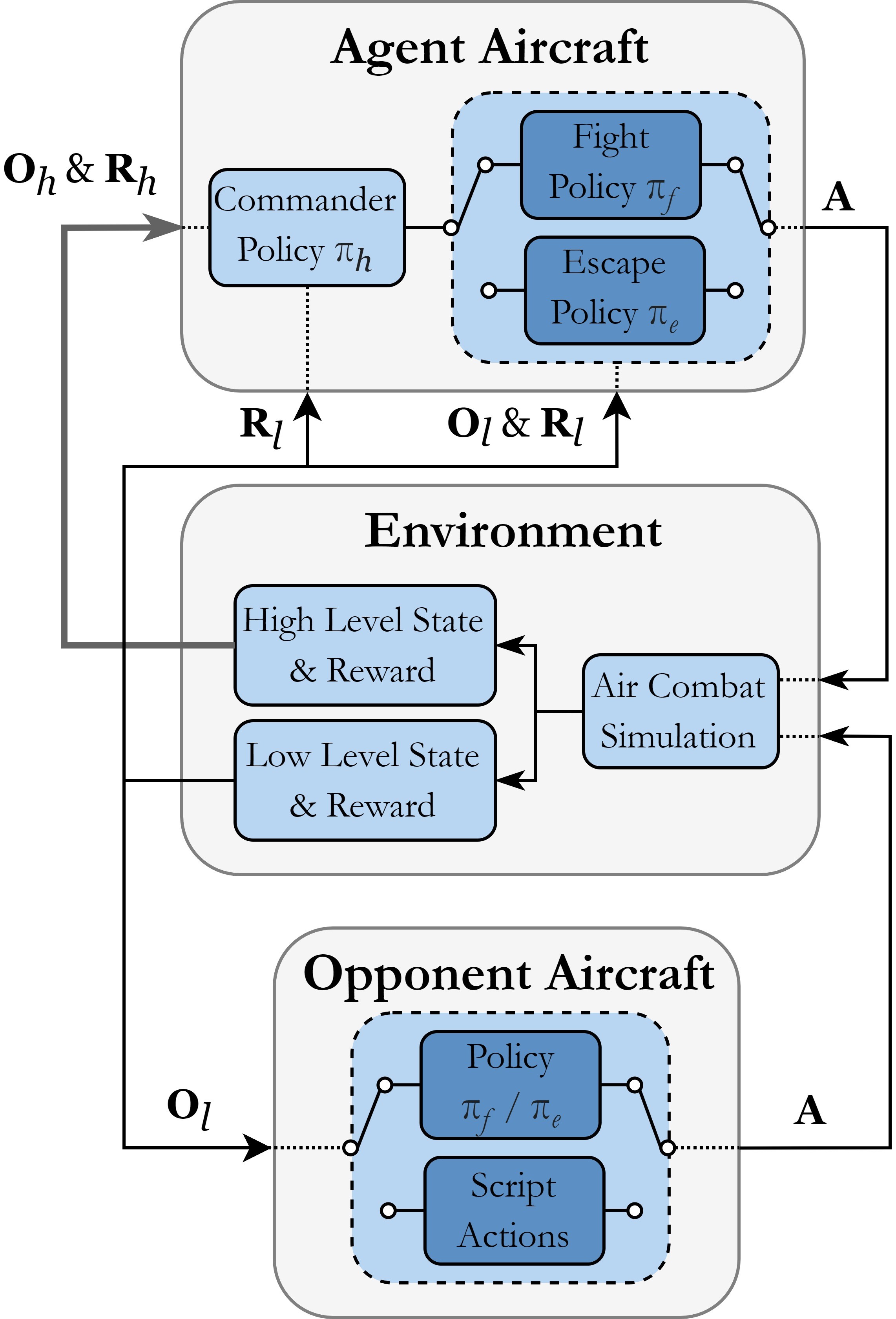}}
\caption{Hierarchical MARL training loop.}
\label{marl_hierarchy_env}
\end{figure}

The overall training loop for our hierarchical MARL algorithm is split into two main stages (Fig~\ref{marl_hierarchy_env}). We first train the low-level policies with observations $O_l$ and rewards $R_l$. In the second stage, low-level policies are fixed (i.e., no learning is done anymore) and serve as options for the commander. $\pi_h$ is then trained with observations $O_h$ and rewards $R_h+R_l$.

The training of low-level policies is done in five levels following a \emph{curriculum learning} scheme. The complexity is increased at each level by making the opponents more competitive. Namely, the opponent behavior of each level is L1 (static), L2 (random), L3 (scripts), L4 (L3 policy), and L5 (policies L1-4). Scripted opponents are programmed to engage the closest agent with $\alpha_{ATA}\approx 0$ and to randomly escape. When training is completed at a level, we transfer the policy to the next level and continue training.

\begin{figure}[htbp]
\centerline{\includegraphics[scale=0.05]{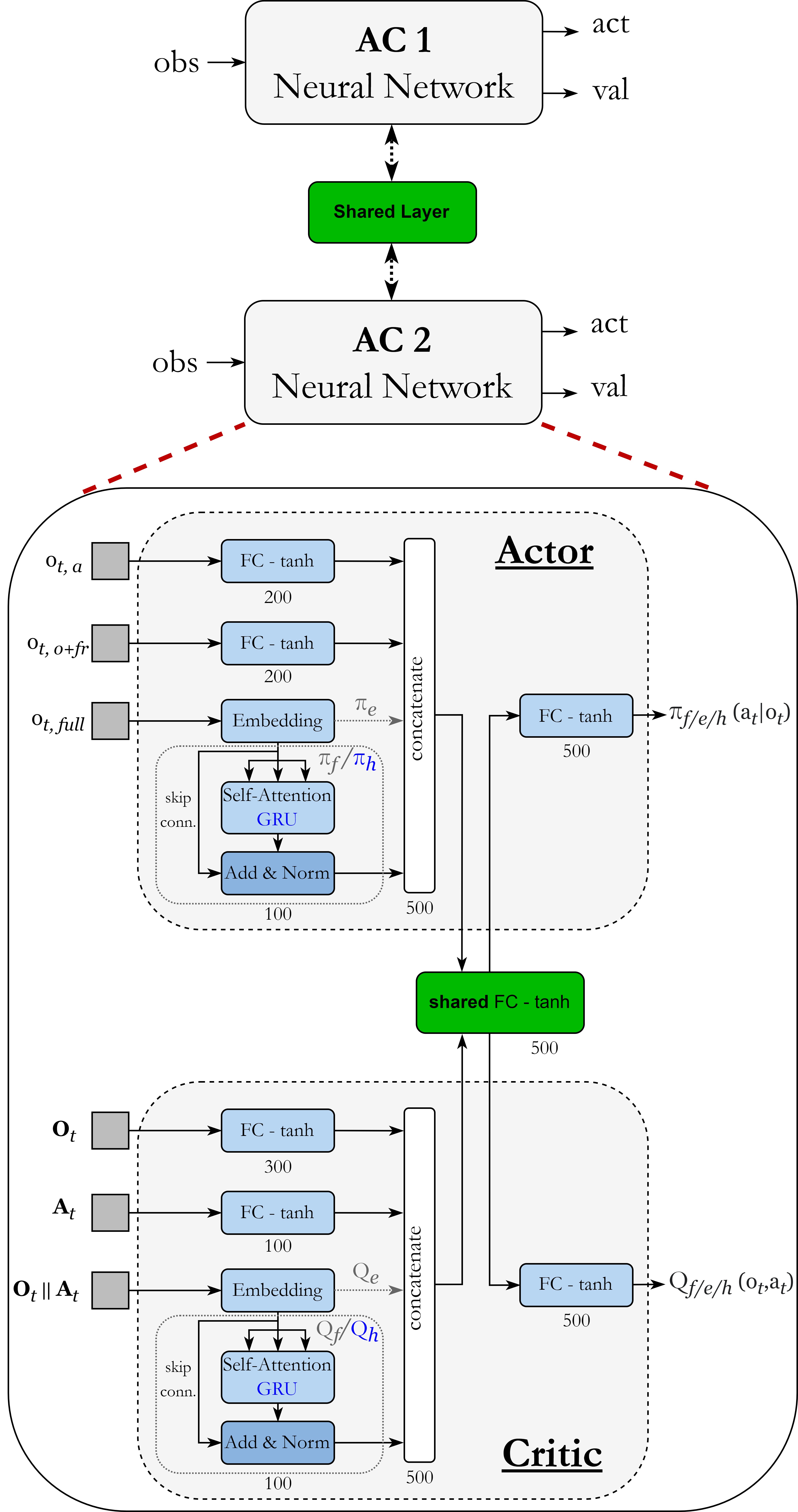}}
\caption{Neural network architecture.}
\label{network_architecture}
\end{figure}

Our neural network is based on Actor-Critic \cite{ac_net} (Fig.~\ref{network_architecture}). Low-level AC1 and AC2 agents have distinct neural network instances (with different input and output dimensions) but share one layer (green box). This layer is further shared between the actor and critic inside the network. Sharing parameters improves agents' coordination \cite{param_sharing}. The architecture modifications are marked for the three policy types. $\pi_f$ uses a \emph{self-attention} (SA) module \cite{attention}, $\pi_h$ a \emph{Gated-Recurrent-Unit} (GRU) module \cite{gru} and $\pi_e$ does not use any of them. The embedding layer is linear with $100$ neurons and $tanh$ activation. High-level commander policy has only one instance for both aircraft types. Since we train the policies with the CTDE scheme, the critic gets the observations of all interacting agents and their actions (global information) as input. Besides parameter sharing, a fully observable critic improves coordination between heterogeneous agents. We update our network parameters using the Actor-Critic approach of Proximal Policy Optimization (PPO) \cite{ppo} (see Alg.~\ref{ppo_algo}). 

\begin{algorithm}[htp!]
\caption{PPO training procedure for $\pi_f$, $\pi_e$ and $\pi_h$}
\begin{algorithmic}[1]
\STATE Set number of episodes $N$, time horizon $T$, batch size $b$ and levels $L$
\STATE Initialize buffer $D$ $\rightarrow$ $\{\}$, policy parameters $\theta$, value function parameters $\phi$
\FOR{level $l=1$ to $L$}
\FOR{episode $n=0$ to $N$}
\STATE Initialize state $S_0$
\FOR{$t=0$ to $T$}
\STATE Get agent actions $A_{t,ag}$ by current policy $\pi_{\theta}$
\STATE Get opp action $A_{t,o}$: script if $l<=3$ else $\pi_{\theta_{l-1}}$
\STATE Execute $(A_{t,ag}, A_{t,o})$, obtain $R_t$ and $S_{t+1}$
\STATE Store $D$ $\leftarrow$ $(S_t, A_{t,ag}, A_{t,o}, R_t, S_{t+1})$
\ENDFOR 
\IF{$|D| >= b$}
\STATE compute advantage estimation $A^{\pi_{\theta}}$ 
\FOR{update iteration $k=1$ to $K$}
\STATE update policy parameters \newline
$\theta_{k+1} = \underset{\theta}{\arg\max} \mathop{\mathbb{E}}_{\tau \sim D}[\sum_{t=0}^{T}[\min( \frac{\pi_{\theta}(a_t|s_t)}{\pi_{\theta_k}(a_t|s_t)}A^{\pi_{\theta_k}},$ $clip(\frac{\pi_{\theta}(a_t|s_t)}{\pi_{\theta_k}(a_t|s_t)}, 1-\epsilon, 1+\epsilon)A^{\pi_{\theta_k}})]]$
\STATE update value function parameters \newline
$\phi_{k+1} = \underset{\phi}{\arg\min} \mathop{\mathbb{E}}_{\tau \sim D} [\sum_{t=0}^{T} (V_{\phi}(s_t)-\hat{R}_t)^2]$
\ENDFOR
\STATE empty buffer $D$ $\rightarrow$ $\{\}$
\STATE set $\theta$ $\leftarrow$ $\theta_{k+1}$
\STATE set $\phi$ $\leftarrow$ $\phi_{k+1}$
\ENDIF
\ENDFOR 
\ENDFOR 
\end{algorithmic}\label{ppo_algo}
\end{algorithm}

\section{Experiments}\label{sec:experiments}

\subsection{Simulation Settings}
We validate our method by simulations.\footnote{Code available at \href{https://github.com/IDSIA/marl}{github.com/IDSIA/marl}.} For this purpose we developed a dedicated 2D (Python) simulation platform to have full control and low inertia. Our platform is lightweight, fast and simulates the dynamics of our aircraft (Sect.~\ref{aircraft_dynamics}). Trajectories of each aircraft can be visualized and a landmark is set at the position where an aircraft got destroyed (Fig.~\ref{trajectories}). Map size and number of interacting aircraft can be specified, highlighting the diversity and scalability properties of our model. We refer to time step $t$ as one simulation round. A simulation episode ends when either the time horizon is reached or there are no alive aircraft of one team. An aircraft is destroyed when getting hit by cannon or rocket or when hitting the map boundary. For each episode, a side of the map (left or right half) is chosen at random for each team, followed by generating random initial positions and headings for each aircraft. Agent training uses the popular libraries \emph{Ray RLlib} and \emph{Pytorch}.

\subsection{Training and Results}
Since we use a shared policy for each aircraft type and the commander, we do not need to restrict our simulations to a fixed number of agents. For every episode, aircraft types are randomly selected, having at least one of each type per group. Map sizes per axis are $30$ km for low-level and $50$ km for high-level policy training. Learning curves showing mean rewards include the performance of all agents. Evaluations are done for $1,000$ episodes. \emph{Win} is when all opponents are destroyed, \emph{loss} if all agents got destroyed and \emph{draw} if at least one agent per team remains alive after the episode ends.

The PPO parameters are kept constant for all training procedures: learning rate actor and critic $lr=0.0001$, discount factor $\gamma=0.95$, clip parameter $\epsilon=0.2$, Adam as optimizer, batch size of $2,000$ for low-level policies and $1,000$ for high-level policy. We train all policies according to Alg.~\ref{ppo_algo} and set the five levels as discussed in Sect.~\ref{method:training_structure}. We have compared the performance of the proposed architecture to a standard RL system, which showed a very poor performance and is left out.

\subsubsection{Low-level policies}
Since each agent can sense only one opponent and one friendly agent at a time, we train our low-level policies in a 2vs2 setting. Since our framework allows an arbitrary number of agents and opponents, we evaluate the performance of each policy type in different combat scenarios. For every new episode, the aircraft ammunition is: $200$ cannon shots and $5$ rockets (AC1). We make the opponents stronger by giving them ammunition of $400$ cannons and $8$ rockets. As the levels increase, we also increase the time horizon of an episode by $\Delta T=50$  starting from $T=200$ on L1.

\begin{figure}[htbp]
\centerline{\includegraphics[scale=0.05]{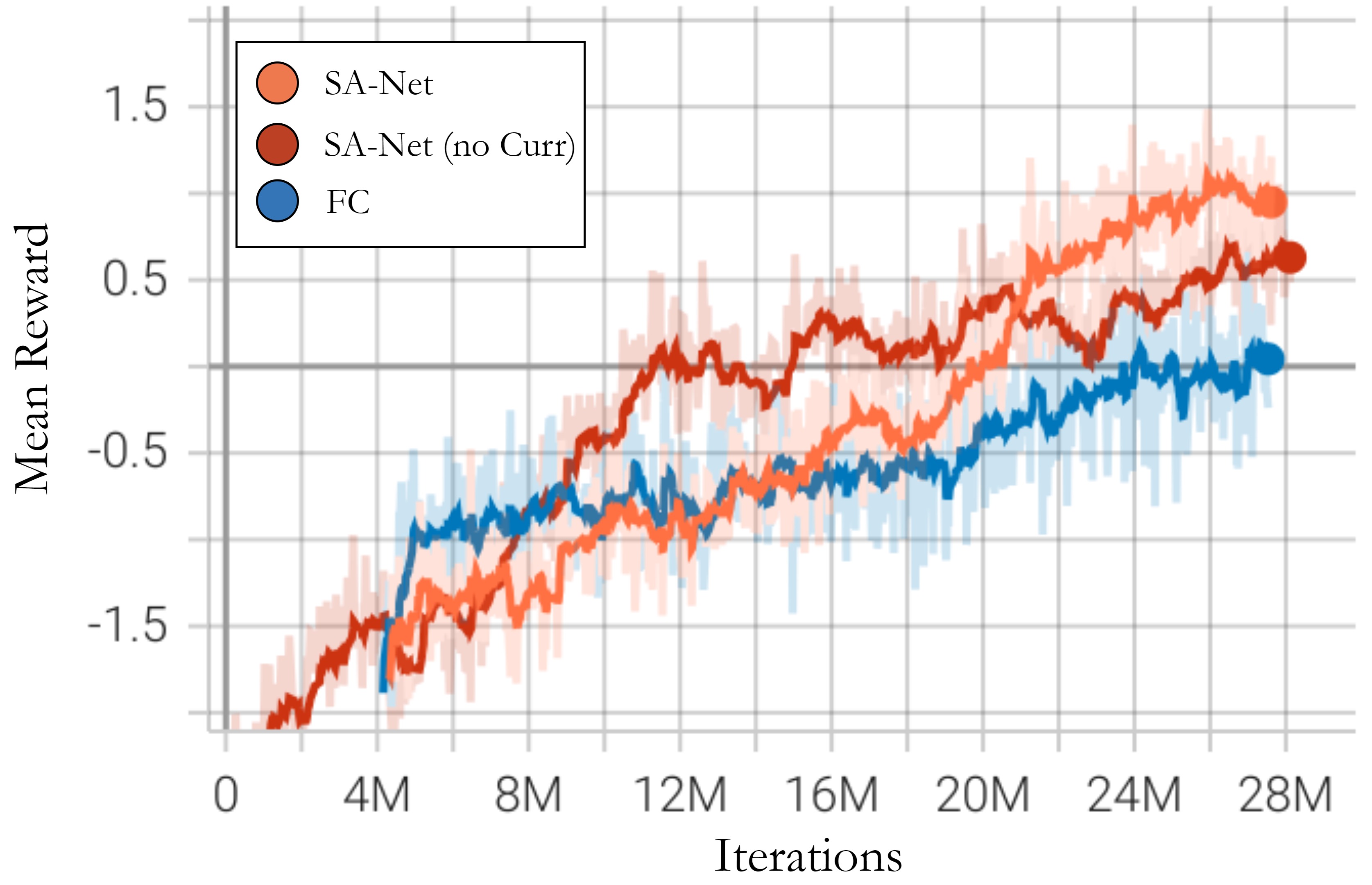}}
\caption{Training performance of $\pi_f$ at L3: SA-Net is our self-attention network (Fig~\ref{network_architecture}), (\emph{no Curr}) is the same network but trained without curriculum, FC is a fully connected network with two layers of $500$ neurons and $tanh$ activation.}
\label{L3_compare}
\end{figure}

Let us first examine the fight policy $\pi_f$. Training starts in L1 and ends in L5. In the latter, the opponents get assigned one of the previous learned fight policies at random for every episode. To highlight the strength of our network architecture, we consider L3, where training is done against script based opponents. The combat behavior of the opponents at this level is the most deterministic, therefore allowing to compare the performance of different architectures (see Fig.~\ref{L3_compare}). 

\begin{figure}[htbp]
\centering
\begin{subfigure}[b]{0.22\textwidth}
\centering
\begin{tikzpicture}
\begin{axis}[
title={\bf {\small Evaluation Level 5}},
ybar,
ymin=0,
ymax=800,
bar width=8,
width=5.25cm,
ymajorgrids,
height=4cm,
xmin=0,
xmax=8,
yticklabels={{\scriptsize 0}, {\scriptsize 0}, {\scriptsize 200}, {\scriptsize 400}, {\scriptsize 600}, {\scriptsize800}},
x tick label style={align=center,text width=1cm},
xticklabels={{\scriptsize k-1}, {\scriptsize k-2}, {\scriptsize d-1}, {\scriptsize d-2}, {\scriptsize fk-1}, {\scriptsize fk-2}, {\scriptsize $ $ Draw}},
xtick={1,2,3,4,5,6,7},
tickwidth=0mm,
axis on top]
\addplot[color=blue, fill=blue!65!lime] coordinates {(1,730) (2,230) (3,200) (4,480) (5,48) (6,39) (7,160)};
\end{axis}
\end{tikzpicture}
\caption{$\pi_f$ 2vs2 statistics}
\label{L5_eval_2vs2}
\end{subfigure}
\hfill
\begin{subfigure}[b]{0.25\textwidth}
\centering
\begin{tikzpicture}
\begin{axis}[
title={\bf {\small Evaluation Fight}},
xbar stacked,
legend entries={{\scriptsize Win},{\scriptsize Loss},{\scriptsize Draw}},
bar width=7,
width=4.7cm,
xmajorgrids,
height=3cm,
xmin=0,
xmax=100,
ytick={0,1,2,3},
yticklabels={{\footnotesize 2vs2},{\footnotesize 3vs3}, {\footnotesize 4vs4}, {\footnotesize 5vs5}},
xtick={0,25,50,75,100},
tickwidth=0mm,
legend style={at={(0.5,-0.35)},anchor=north,legend columns=-1},
xticklabels={{\scriptsize $0\%$},{\scriptsize $25\%$}, {\scriptsize $50\%$},{\scriptsize $75\%$},{\scriptsize $100\%$}},
axis on top]
\addplot [color=blue, fill=blue!65!lime] coordinates
{(71,0) (65,1) (56,2) (47,3)};
\addplot [color=red, fill=red!60!orange] coordinates
{(10,0) (11,1) (14,2) (16,3)};
\addplot [color=gray, fill=gray!60!white] coordinates
{(19,0) (24,1) (30,2) (37,3)};
\end{axis}
\end{tikzpicture}
\caption{$\pi_f$ evaluation scenarios}
\label{L5_eval_scenarios}
\end{subfigure}
\caption{Evaluation of $\pi_f$ after finishing L5 training. Destroying an opponent is abbreviated with \emph{k}, getting destroyed with \emph{d} and \emph{fk} indicates ``friendly'' kills. The attached numbers indicate the aircraft types, e.g. $k-1$ kill by AC1, $d-2$ AC2 destroyed.}
\label{L5_eval}
\end{figure}
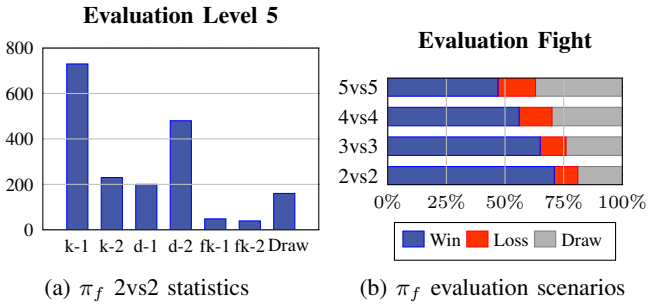

\begin{figure*}[htbp]
\centering
\begin{subfigure}[b]{0.32\textwidth}
\centering
\includegraphics[width=\textwidth]{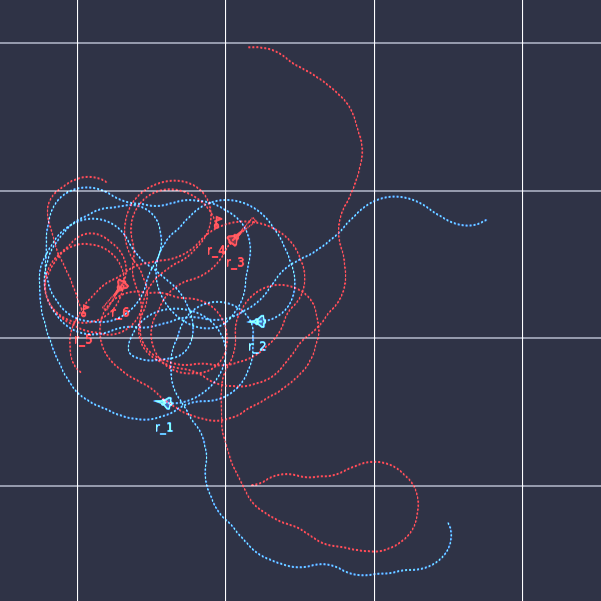}
\caption{Commander}
\label{traj_hier}
\end{subfigure}
\hfill
\begin{subfigure}[b]{0.32\textwidth}
\centering
\includegraphics[width=\textwidth]{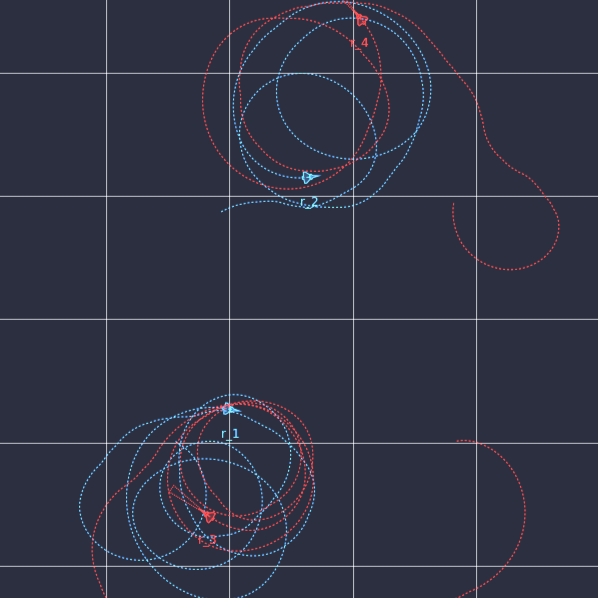}
\caption{Fight}
\label{traj_fight}
\end{subfigure}
\hfill
\begin{subfigure}[b]{0.32\textwidth}
\centering
\includegraphics[width=\textwidth]{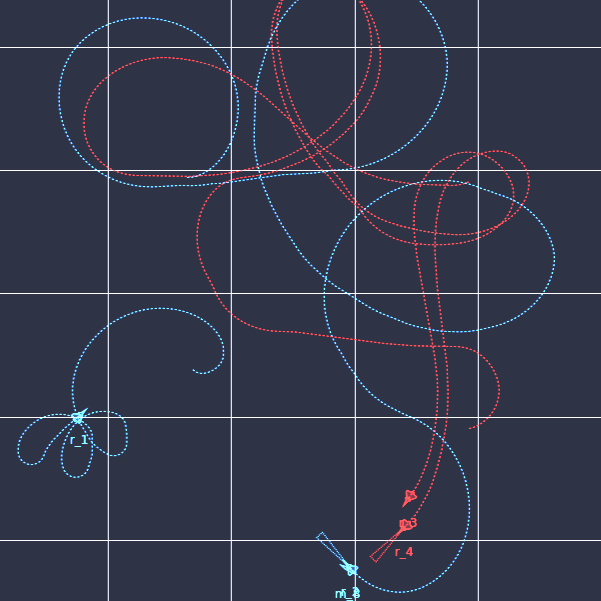}
\caption{Escape}
\label{traj_esc}
\end{subfigure}
\caption{Different policy trajectories in the simulation platform.}
\label{trajectories}
\end{figure*}

We evaluate the performance of our agents when training has completed (Fig.~\ref{L5_eval}). We deploy every agent with $\pi_f$ of L5 and every opponent with $\pi_f$ of L4. The combat skills of AC1 clearly surpass those of AC2, which is most likely due to rockets as further equipment and having more agile dynamics. We infer that our agents could further improve their combat performance during L5 training and are able to combat in scenarios up to 5vs5, even though training was conducted in a 2vs2 scheme. However, as the number of aircraft increases, the portion of draws also rises. This observation may suggest that the low-level policy has reached its peak learning capacity. An example of a fight scenario is shown in Fig~\ref{traj_fight}, showing the circular trajectories to reach the tail of the opponents.

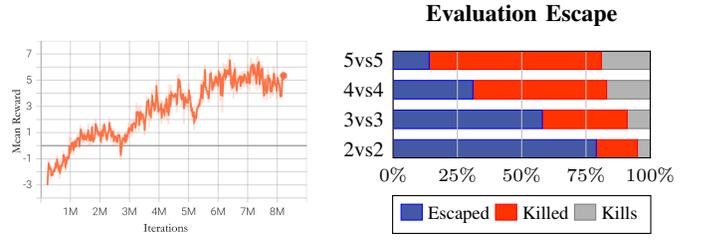
\begin{figure}[htbp]
\centering
\begin{subfigure}[b]{0.22\textwidth}
\centering
\includegraphics[width=\textwidth]{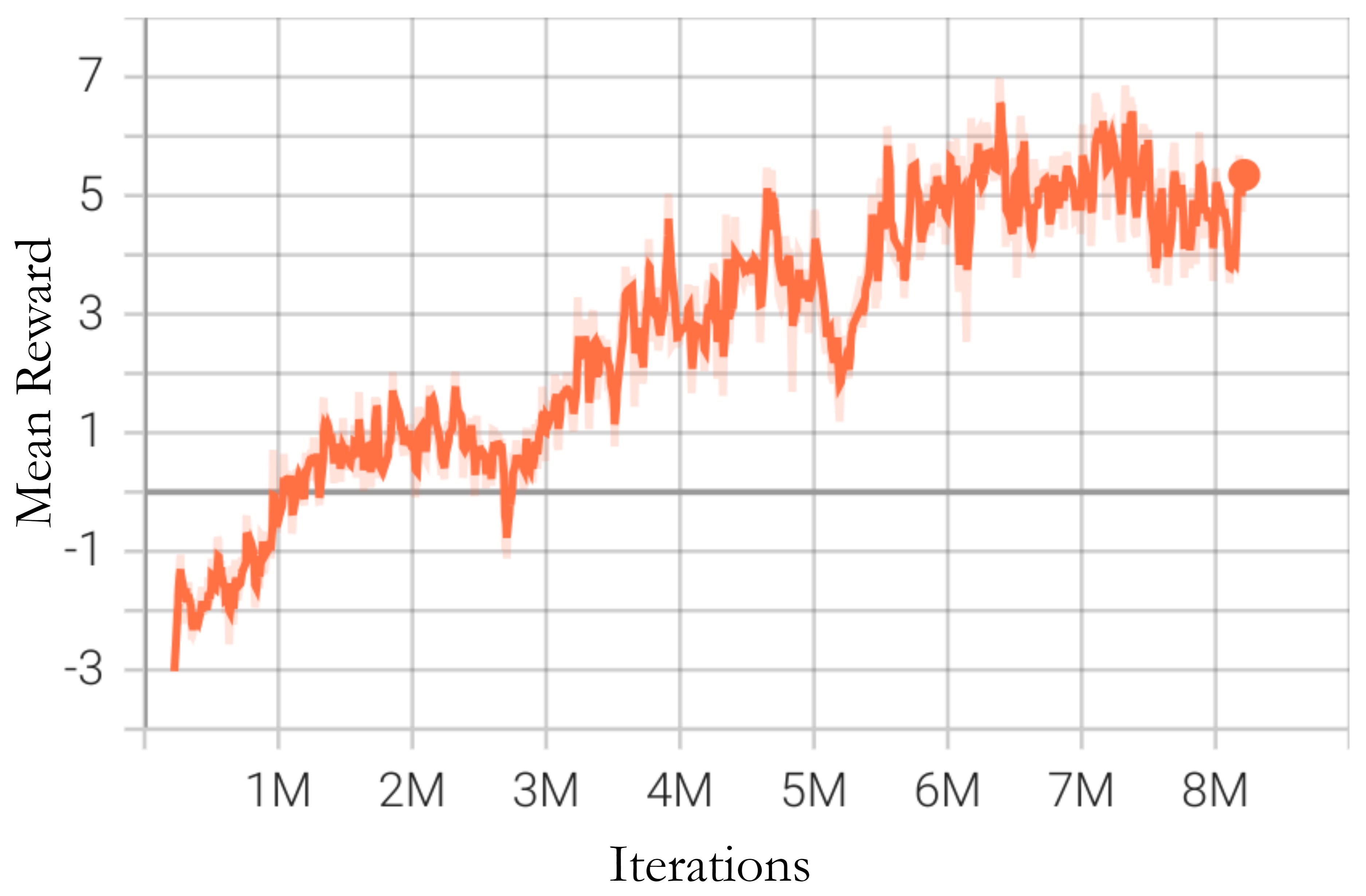}
\caption{$\pi_e$ training performance}
\label{esc_reward}
\end{subfigure}
\hfill
\begin{subfigure}[b]{0.25\textwidth}
\centering
\begin{tikzpicture}
\begin{axis}[
title={\bf {\small Evaluation Escape}},
xbar stacked,
legend entries={{\scriptsize Escaped},{\scriptsize Killed},{\scriptsize Kills}},
bar width=7,
width=5cm,
xmajorgrids,
height=3cm,
xmin=0,
xmax=100,
ytick={0,1,2,3},
yticklabels={{\footnotesize 2vs2},{\footnotesize 3vs3}, {\footnotesize 4vs4}, {\footnotesize 5vs5}},
xtick={0,25,50,75,100},
tickwidth=0mm,
legend style={at={(0.5,-0.35)},anchor=north,legend columns=-1},
xticklabels={{\scriptsize $0\%$},{\scriptsize $25\%$}, {\scriptsize $50\%$},{\scriptsize $75\%$},{\scriptsize $100\%$}},
axis on top]
\addplot [color=blue, fill=blue!65!lime] coordinates
{(79,0) (58,1) (31,2) (14,3)};
\addplot [color=red, fill=red!60!orange] coordinates
{(16,0) (33,1) (52,2) (67,3)};
\addplot [color=gray, fill=gray!60!white] coordinates
{(5,0) (9,1) (17,2) (19,3)};
\end{axis}
\end{tikzpicture}
\caption{$\pi_e$ evaluation scenarios}
\label{esc_scenarios}
\end{subfigure}
\caption{Training performance of $\pi_e$ at L3 (a) and different combat scenarios (b). \emph{Escaped} means no agent got killed (including going out of boundary), \emph{killed} is when at least one agent got killed, \emph{kills} when at least one opponent got killed.}\label{esc_training}
\end{figure}

The escape policy $\pi_e$ has the purpose of fleeing from opponents. We consider only L3 for training and show the training results in Fig.~\ref{esc_reward}. Agents can still fire and destroy opponents, but training results indicate the correct behavior by exploiting the escaping reward more than the killing reward. As the number of aircraft increases in evaluation, the agents are less capable of fleeing successfully from opponents (Fig.~\ref{esc_scenarios}). Fleeing trajectories are visualized in Fig~\ref{traj_esc}.

\begin{algorithm}[htp!]
\caption{Simulation procedure for commander policy $\pi_h$}
\begin{algorithmic}[1]
\STATE Set high-level time horizon $T_h=40$, low-level time horizon $T=10$ and max number of episodes $N$
\FOR{episode $n=0$ to $N$}
\STATE Sample a 3vs3 scenario
\FOR{$t_h=0$ to $T_h$}
\STATE Get commander actions for each agent $A_{h,t_h}$
\STATE Activate corresponding low-level policies $\pi_f$ or $\pi_e$
\FOR{$t=0$ to $T$}
\STATE Execute $\pi_f$ or $\pi_e$ for each agent and opponent
\IF{$t\geq 10$ or event}
\STATE \textbf{break}
\ENDIF
\ENDFOR 
\STATE Get reward and next state
\ENDFOR 
\ENDFOR 
\end{algorithmic}
\label{hier_algo}
\end{algorithm}

\subsubsection{High-level policy}
The purpose of the commander policy $\pi_h$ is to provide strategic commands (attack or escape). Since $\pi_h$ can observe three agents and three opponents per time, we do a 3vs3 combat training. Agents and opponents are equipped either with $\pi_f$ of L5 or $\pi_e$. Opponents are mainly set to fight and randomly to escape. Ammunition is set to $300$ cannons and $8$ rockets. We include the low-level policies as part of the environment (low-level dynamics in Fig.~\ref{hierarchy_policy}). We run experiments according to the procedure in Alg.~\ref{hier_algo}. The commander gets invoked dynamically on events or when a low-level horizon is reached. Events are characterized as:
\begin{itemize}
    \item any aircraft got destroyed (by shooting or hitting map boundary);
    \item an agent approaches the map boundary ($d<6km$);
    \item an agent \emph{or} an opponent is in favorable situation as described in Eq.~\eqref{fav_sit_reward};
    \item two opponents are close and face an agent ($d<5km$, $\alpha_{ATA,a}<30^\circ$).
\end{itemize}

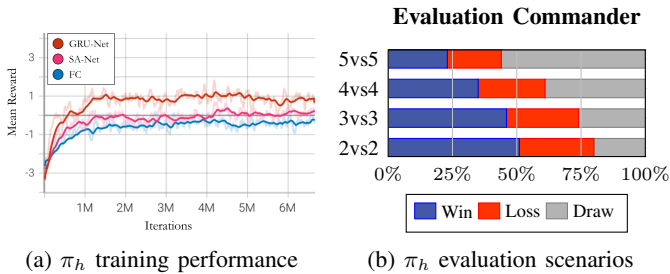
\begin{figure}[htbp]
\centering
\begin{subfigure}[b]{0.23\textwidth}
\centering
\includegraphics[width=\textwidth]{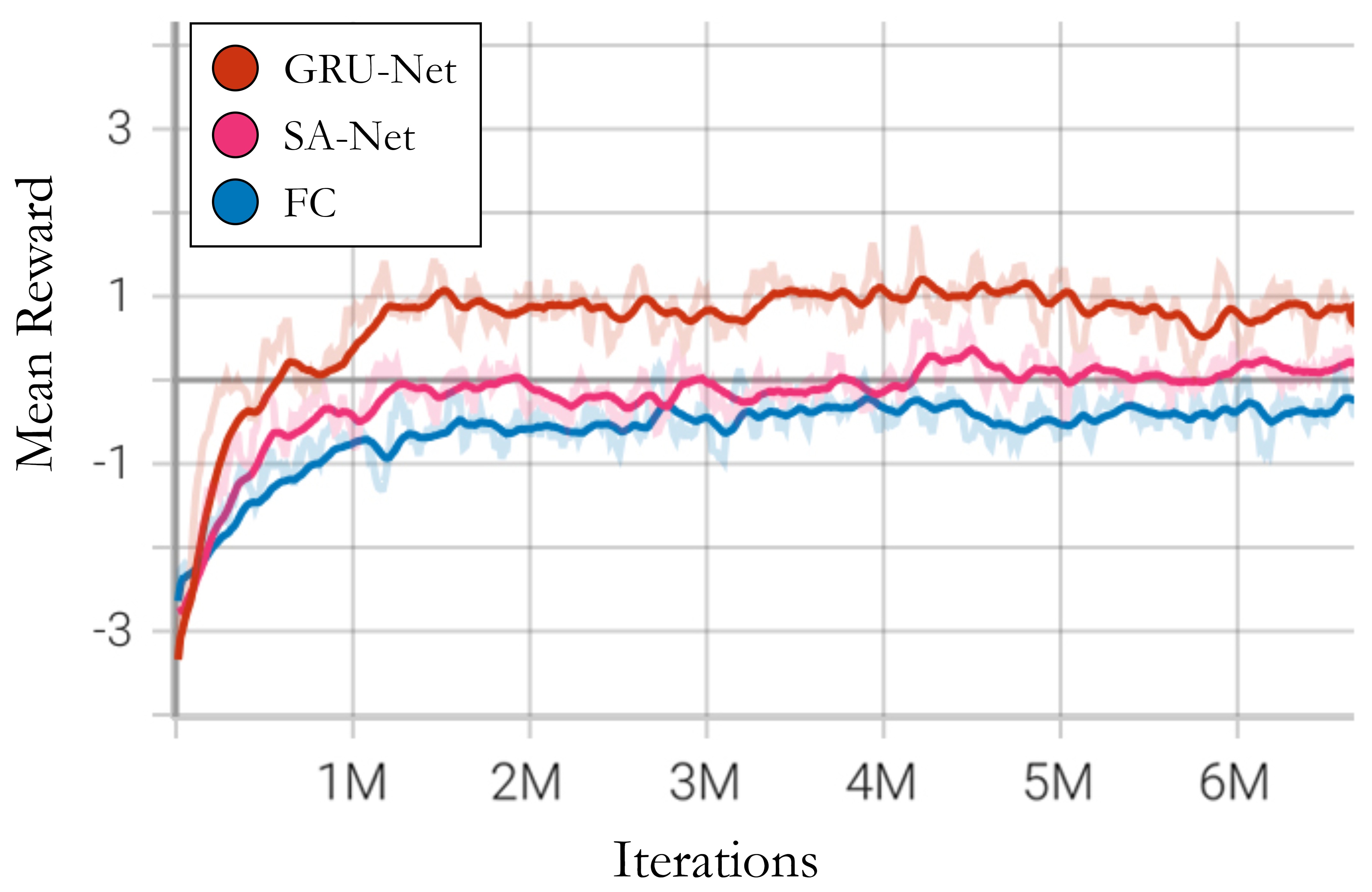}
\caption{$\pi_h$ training performance}
\label{hier_comp_training}
\end{subfigure}
\hfill
\begin{subfigure}[b]{0.25\textwidth}
\centering
\begin{tikzpicture}
\begin{axis}[
title={\bf {\small Evaluation Commander}},
xbar stacked,
legend entries={{\scriptsize Win},{\scriptsize Loss},{\scriptsize Draw}},
bar width=7,
width=5cm,
xmajorgrids,
height=3cm,
xmin=0,
xmax=100,
ytick={0,1,2,3},
yticklabels={{\footnotesize 2vs2},{\footnotesize 3vs3}, {\footnotesize 4vs4}, {\footnotesize 5vs5}},
xtick={0,25,50,75,100},
tickwidth=0mm,
legend style={at={(0.5,-0.35)},anchor=north,legend columns=-1},
xticklabels={{\scriptsize $0\%$},{\scriptsize $25\%$}, {\scriptsize $50\%$},{\scriptsize $75\%$},{\scriptsize $100\%$}},
axis on top]
\addplot [color=blue, fill=blue!65!lime] coordinates
{(51,0) (46,1) (35,2) (23,3)};
\addplot [color=red, fill=red!60!orange] coordinates
{(29,0) (28,1) (26,2) (21,3)};
\addplot [color=gray, fill=gray!60!white] coordinates
{(20,0) (26,1) (39,2) (56,3)};
\end{axis}
\end{tikzpicture}
\caption{$\pi_h$ evaluation scenarios}
\label{hier_eval_scenarios}
\end{subfigure}
\caption{Training performance of $\pi_h$ for different architectures. SA-Net is the network used to train $\pi_f$, GRU-Net is the network in Sect.~\ref{method:training_structure}, FC is a fully connected network with two layers of 500 neurons and $tanh$ activation.
\label{hier_scenarios}}
\end{figure}

Training results are in Fig~\ref{hier_scenarios}. The learning curves of all models quickly saturate (Fig~\ref{hier_comp_training}), but the GRU module improves the result by storing the last state. We choose this model as our commander $\pi_h$ and evaluate the performance. We infer that the commander $\pi_h$ improves combat performance for small team sizes. However, as the number of aircraft increases in evaluation scenarios (Fig~\ref{hier_eval_scenarios}), the result tends to a draw and an equal win-to-loss ratio, which we could also expect when no commander would be involved, since both teams are equally equipped (except for number of AC1 and AC2 per team). The reason for this is the partial observability of only three opponents around an agent. Another reason might be the stochasticity involved, where an opponent suddenly might switch from fight to escape, affecting the coordination of the commander. A further aspect for not achieving superiority in larger team sizes is the number of AC1 and AC2 per team, since AC1 has stronger combat performance as shown in Fig~\ref{L5_eval_2vs2}. An example of a 2vs4 combat scenario with the commander involved is shown in Fig~\ref{traj_hier}, where two opponents could successfully be destroyed within the time horizon.

\section{Conclusion}\label{sec:conclusions}
We presented a hierarchical, heterogeneous, multi-agent reinforcement learning procedure for air-to-air combat maneuvering. The key ideas are using curriculum learning, fictitious self-play and sophisticated neural networks. The empirical validation shows the promising potential of our design. Our agents can effectively engage in air-to-air combat with solid resilience, while the commander has difficulties in successfully coordinating larg team configurations. In future work, we intend further to improve the hierarchical structure for better tactical decisions, regardless of the team size. We also plan to incorporate a dedicated communication mechanism as well as to switch to 3D aircraft models for a more realistic environment and accurate aircraft dynamics.

\bibliographystyle{IEEEtran}
\bibliography{biblio}
\end{document}